%% file: main.tex
\documentclass[letterpaper,10pt,conference]{ieeeconf}  
\IEEEoverridecommandlockouts                              
\usepackage{algorithm}  
\usepackage{algpseudocode}  
\usepackage{amsmath} 
\usepackage{amssymb}  
\usepackage{booktabs} 
\usepackage{epsfig} 
\usepackage{graphics} 
\usepackage{mathptmx} 
\usepackage{subcaption} 
\usepackage{times} 
\usepackage{caption}
\title{\LARGE \bf
    On Learning Closed-Loop Probabilistic Multi-Agent Simulator
}

\author{%
Juanwu~Lu$^{1}$, Rohit~Gupta$^{2}$, Ahmadreza Moradipari$^{2}$, Kyungtae~Han$^{2}$, Ruqi~Zhang$^{1}$, and Ziran~Wang$^{1}$
\thanks{$^{1}$J. Lu, Z. Wang, and R. Zhang are with Purdue University, West Lafayette, IN 47907, USA}%
\thanks{$^{2}$R. Gupta, A. Moradipari, and K. Han are with Toyota InfoTech Labs, Mountain View, CA 94043, USA}%
\thanks{Corresponding author: J. Lu, email: {\tt\small juanwu@purdue.edu.}}%
}

\begin{document}

\maketitle
\thispagestyle{empty}
\pagestyle{empty}

\input{sections/0_abstract}
\input{sections/1_intro}
\input{sections/3_method}

\input{sections/4_experiment}

\input{sections/5_conclusion}

\input{main.bbl}
\end{document}

%% file: sections/0_abstract.tex
\begin{abstract}
    The rapid iteration of autonomous vehicle (AV) deployments leads to increasing needs for building realistic and scalable multi-agent traffic simulators for efficient evaluation. Recent advances in this area focus on closed-loop simulators that enable generating diverse and interactive scenarios. This paper introduces \textit{Neural Interactive Agents} (NIVA), a probabilistic framework for multi-agent simulation driven by a hierarchical Bayesian model that enables closed-loop, observation-conditioned simulation through autoregressive sampling from a latent, finite mixture of Gaussian distributions. We demonstrate how NIVA unifies preexisting sequence-to-sequence trajectory prediction models and emerging closed-loop simulation models trained on Next-token Prediction (NTP) from a Bayesian inference perspective. Experiments on the Waymo Open Motion Dataset demonstrate that NIVA attains competitive performance compared to the existing method while providing embellishing control over intentions and driving styles. 
\end{abstract}

%% file: sections/1_intro.tex
\section{Introduction}
\label{sec: introduction}

\subsection{Background}
\label{subsec: background}

Despite the rapid advancement of autonomous driving technologies, we expect a prolonged transitional period in which human-driven vehicles (HDVs) and autonomous vehicles (AVs) coexist~\cite{fagnant2015preparing}. In such mixed traffic, human behavior is often heterogeneous and unpredictable, while AVs are required to plan and execute optimal trajectories based on well-defined criteria. These complex, multi-agent interactions are critical to traffic safety and demand robust evaluation and prototyping tools for AV development.

Fleet-based empirical testing remains the industry standard but is expensive, time-consuming, and lacks reproducibility. \emph{Log-replay} datasets~\cite{Ettinger_2021_ICCV, Chang_2019_CVPR} address these issues by replaying recorded motions of surrounding agents during test drives. However, these datasets do not allow log-replay agents to respond to the evolving behavior of the AV, leading to unrealistic and ineffective evaluations~\cite{9561666}. Consequently, there is growing interest in interactive, closed-loop traffic simulation~\cite{NEURIPS2023_b96ce67b}.

Traditional simulation methods—often deterministic and open-loop—fail to capture the inherent uncertainty, multimodality, and dynamic interactivity of real-world traffic. This has motivated a wave of learning-based approaches that model traffic as a stochastic process. Recent progress in closed-loop simulation with learning-based models has achieved state-of-the-art accuracy in multi-agent trajectory prediction. However, many of these models act as black boxes, offering limited interpretation and control over the underlying dynamics. This opacity makes it difficult to diagnose fidelity issues, refine behavior conditionally, or adapt models to new scenarios~\cite{lu2022generalizability}. To address these limitations, recent work has incorporated probabilistic modeling~\cite{Suo_2021_CVPR, Lu_2024_CVPR} and explored ways to represent social interactions and agent preferences during future trajectory generation~\cite{10172005}.

\begin{figure}
    \centering
    \includegraphics[width=\linewidth]{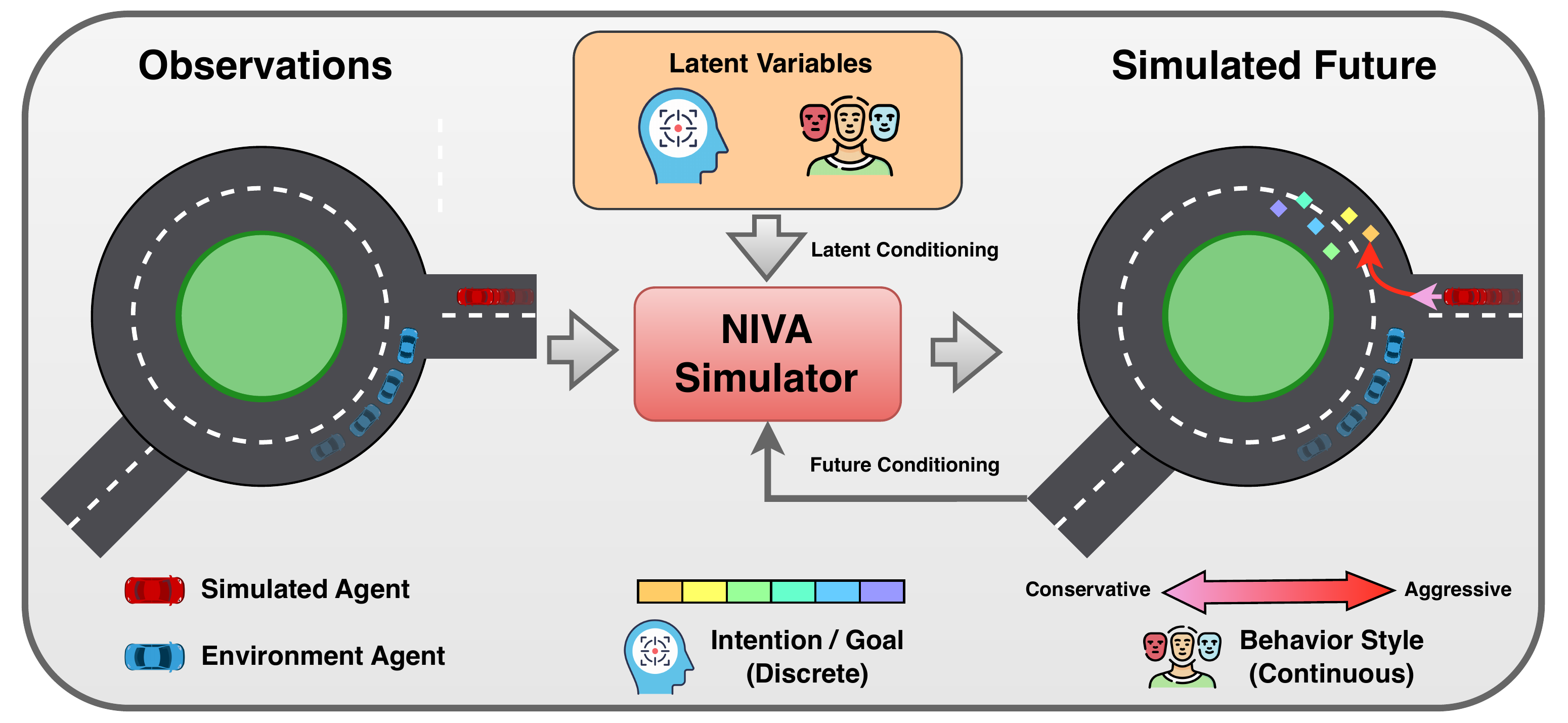}
    \caption{\textbf{Overview of our simulation framework.} NIVA models disentangled latent factors governing behavior styles, discrete intentions, and next-step dynamics for multi-agent interaction.}
    \label{fig: illustration}
\end{figure}

\subsection{Related Works}
\label{subsec: related-works}

Early efforts focused on open-loop modeling using recorded trajectories~\cite{Gao_2020_CVPR, NEURIPS2022_2ab47c96}, which lacked the ability to account for dynamic and interactive behaviors. Recent benchmarks~\cite{NEURIPS2023_b96ce67b} have exposed these limitations, driving interest toward closed-loop simulation frameworks. The key distinction lies in conditioning: open-loop models predict future trajectories based solely on past observations, while closed-loop models continuously condition actions on evolving states as the simulation unfolds.

Closed-loop modeling poses new challenges in recursive generation and scenario consistency. Early works addressed this using Markovian or recurrent models~\cite{9561666, Suo_2021_CVPR}. More recently, Transformer-based architectures have gained traction due to their scalability and success in autoregressive generation~\cite{zhang2024trafficbots, zhou2024behaviorgptsmartagentsimulation, NEURIPS2024_cef5c8de}. Some works even explore adapting pre-trained Large Language Models (LLMs) for scenario generation~\cite{10629039}.

In parallel, generative probabilistic models have proven effective in simulating complex stochastic systems, from weather forecasting~\cite{doi:10.1073/pnas.1810286115} to particle interactions~\cite{de2017learning}. Unlike deterministic simulators, probabilistic models explicitly learn the distribution of observations, allowing them to capture multi-modality, constraints, and uncertainty. Mixtures of Gaussians are commonly used to model the multimodal nature of human behavior~\cite{NEURIPS2022_2ab47c96, 10398503}, while others have used probabilistic modeling to improve generalization~\cite{pmlr-v238-lu24a}, quantify uncertainty~\cite{Lu_2024_CVPR}, and enable controllable scenario generation~\cite{10172005}.

Despite these achievements, two key limitations remain underexplored. First, open-loop and closed-loop simulation models are typically treated as separate modeling paradigms---one-shot generation vs.\ autoregressive generation---without a unified probabilistic framework that bridges both. Meanwhile, existing models often function as opaque black boxes, limiting insight and control over the underlying behavioral dynamics. This impedes interpretability, scenario debugging, and adaptation to novel environments.

\subsection{Contributions}
To address these gaps, we propose \textit{Neural Interactive Agents} (NIVA), a unified probabilistic framework for multi-agent traffic simulation. Our approach reformulates the simulation task as learning a factorized generative process over behavior styles, discrete intentions, and agent dynamics.

As illustrated in Fig.~\ref{fig: illustration}, NIVA is a hierarchical probabilistic model that disentangles latent variables corresponding to (i) continuous \textbf{behavior styles}, (ii) discrete \textbf{intentions}, and (iii) the high-dimensional representation of the \textbf{next-step actions} constrained by the surrounding traffic. At its core is a decoder-only Transformer with \emph{adaptive Layer Normalization}, where the normalization parameters are dynamically modulated by the latent style and intention variables. This design enables NIVA to generate multi-agent interactions conditioned on interpretable behavioral factors while unifying open-loop and closed-loop generation via hierarchical generative process.

We validate our framework on the Waymo Open Motion Dataset~\cite{Ettinger_2021_ICCV}, demonstrating that NIVA achieves competitive performance with state-of-the-art baselines, while learning semantically meaningful, disentangled latent spaces that can be sampled for diverse and realistic behavior generation.

%% file: sections/3_method.tex
\begin{figure*}[t!]
    \centering
    \includegraphics[width=\linewidth]{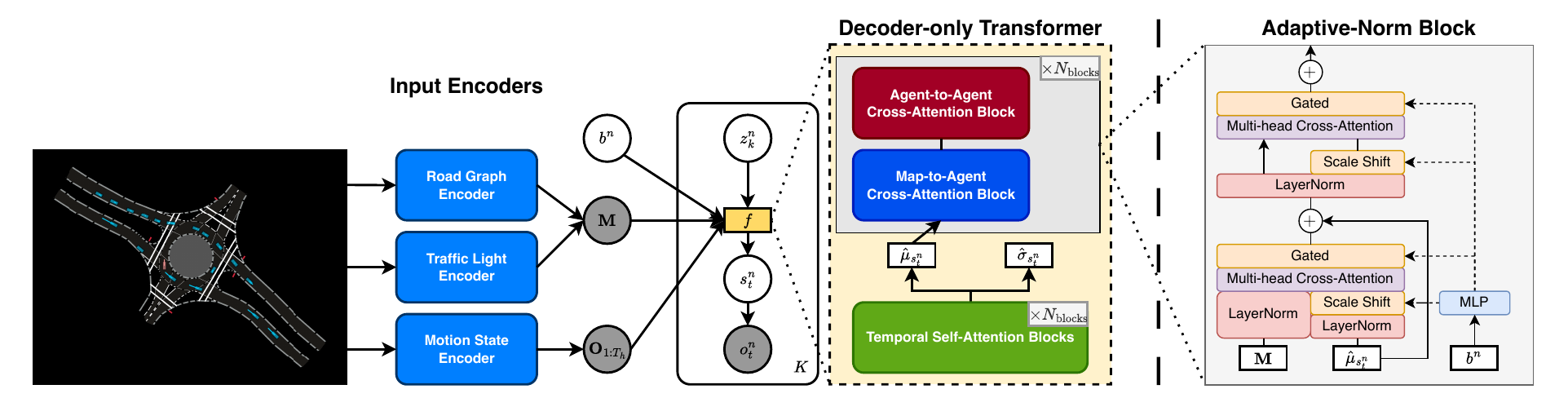}
    \caption{\textbf{The proposed Neural Interactive Agents (NIVA) architecture} has three encoders for map geometry, traffic signals, and trajectory observations. Encoded features are inputs for a three-level hierarchical Bayesian model. The Adaptive-Norm Transformer in the transition model has attention layers learning correlations between agent and map and among agents, and adaptive layer normalization to condition dynamics on behavior style and intentions.}
    \label{fig: architecture}
\end{figure*}

\section{Problem Statement}
\label{sec: problem}

\begin{table}[!t]
    \centering
    \caption{Table of Notations}
    \begin{tabular}{@{}c|l@{}}
        \toprule
         \textbf{Notation} & \textbf{Explanation}  \\
         \midrule
         $N$    & Number of simulated traffic participants. \\
         $T_{h}$    & Number of time steps observed in the history.     \\
         $T_{p}$    & Number of time steps to simulate in the future.   \\
         $\boldsymbol{o}_{t}^{n}$   & Motion state of $n$-th agent at time step $t$.    \\
         $\mathbf{O}_{t}$   & A collection of all motion state observation at time step $t$. \\
         $\boldsymbol{m}_{j}$   & $j$-th map feature. \\
         $\mathbf{M}$   & A collection of map features. \\
         $\mathbf{b}$   & A continuous random variable encoding driving styles.  \\
         $\mathbf{z}$   & A discrete random variable encoding intentions. \\
         $\boldsymbol{s}_{t}^{n}$   & Latent motion state representation of $n$-th agent at time $t$. \\
         $\mathbf{S}_{t}$   & The collection of all latent motion state s at time $t$. \\
         $p(\cdot)$     & Probability density in generative process. \\
         $q(\cdot)$     & Probability density in variational distributions. \\
         $\mathcal{N}(\mu,\sigma^{2}\mathbf{I})$  & Gaussian distribution with diagonal covariance matrix. \\
         \bottomrule
    \end{tabular}
    \label{tab:notation}
\end{table}

This paper proposes a probabilistic multi-agent traffic simulator. We focus on simulating realistic driver behaviors conditioned on observations from a fixed perception system. For readability, Table~\ref{tab:notation} lists the notations used in this paper.

\paragraph{Nomenclature} A traffic simulator is a model that allows the generation of scenarios described by the road infrastructure and the motion of traffic participants. We define the following terms:
\begin{itemize}
    \item A \emph{scenario} consists of observations of $N$ traffic participants, where we have observed $T_{h}$ timesteps and would like to simulate for the consecutive $T_{p}$ timesteps.
    \item An \emph{observation} at the time step $t$, denoted by the notation $\mathbf{O}_{t}=\left\{o_{t}^{1},\ldots,o_{t}^{n},\ldots,o_{t}^{N}\right\}$, is the set of vectors that describes the motion states of the agent $n=1, \ldots, N$.
    \item The \emph{map features}, denoted arbitrarily by $\mathbf{M}$, is the collection of observations about the surroundings, including road geometry, obstacles, and traffic signals.
    \item An \emph{agent} refers to a controllable traffic participant in the scenario, either an HDV or an AV.
\end{itemize}

\paragraph{Problem Formulation} Following the existing works~\cite{Suo_2021_CVPR, NEURIPS2023_b96ce67b}, a traffic simulator unrolls scenarios sequentially by conditioning the motion states of agents at the next time step on the history. To learn a probabilistic simulator parameterized by $\theta$, the objective is to solve for the optimal parameters $\theta^{\ast}$ on a parameter space $\Theta$ that maximizes the log-likelihood of the observations in the training dataset, \emph{i.e.},
\begin{equation}
    \label{eq: objective}
    \theta^{\ast}\gets\underset{\mathbf{\theta}\in\Theta}{\text{argmax}}\mathbb{E}_{\left\{\mathbf{O},\mathbf{M}\right\}\sim\mathcal{D}_{\text{train}}}\sum_{n=1}^{N}\sum_{t=1}^{T}\log p(o_{t}^{n}\mid\mathbf{O}_{<t},\mathbf{M};\theta)
\end{equation}

An obvious limitation of the objective specified by equation~(\ref{eq: objective}) is that it does not explicitly represent the driving styles of different individuals. We address this limitation by introducing a continuous latent variable $\mathbf{b}$, where each agent is explicitly associated with $b^{n}$ to capture the variability in behavior. In addition, since the destination of each individual is generally unobservable, agents can have multiple plausible future movements, each corresponding to a different intention (\emph{e.g., turn left, go straight, or turn right}) when conditioned on the same history observation $\mathbf{O}_{1:T_{h}}$. To account for this phenomenon, we introduce a discrete random variable $\mathbf{z}$, where $z_{k}^{n}$ encodes the $k$-th intention of the $n$-th agent. Intentions are mutually exclusive. The new objective with factorized probability density writes:
\begin{equation}
    \label{eq: model-objective}
    \underset{\mathbf{\theta}\in\Theta}{\text{argmax}}\mathbb{E}_{\mathcal{D}_{\text{train}}}\sum_{n=1}^{N}p(b^{n})\sum\limits_{k=1}^{K}p(z^{n}_{k})\sum_{t=1}^{T}\log p(o_{t}^{n}\mid\mathbf{o}_{<t},\mathbf{M},z^{n}_{k},b^{n};\mathbf{\theta}),
\end{equation}
where $K$ is the number of intentions. In the next section, we extend the objective to a Bayesian inference problem and propose the NIVA model with a recognition model that approximates the posterior distribution of the latent variables. The extension allows us to learn a generative process for sampling diverse and realistic traffic scenarios and a recognition model that supports efficient conditioning on future observations.

\section{Neural Interactive Agents}
\label{sec: method}

\subsection{Probabilistic Model}
\label{subsec: probabilistic}

To bridge the gap between open-loop and closed-loop trajectory generation paradigms, NIVA assumes the following generative process for a multi-agent traffic scenario:
\begin{enumerate}
    \item Agents arrive by the Poisson process: $N\sim\text{Poisson}(\xi)$.
    \item At the beginning of the simulation, the individual behavior style features are i.i.d. samples from the standard Gaussian prior $b^{n}\overset{\text{i.i.d.}}{\sim}\mathcal{N}(0,\mathbf{I}),n=1, \ldots, N$.
    \item For each agent $n=1,\ldots,N$:
    \begin{enumerate}
        \item Choose its long-term intention from a categorical prior distribution $z^{n}\in\left\{1,\ldots,K\right\}\sim\text{Cat}(\pi)$.
        \item Conditioned on the history observation, the simulator samples high-dimensional latent features representing the future trajectory from an open-loop conditional prior distribution
        \begin{equation}
            s_{T_{h}+t}^{n}\sim{
            p\left(s_{t}^{n}
            \mid{\mathbf{O}_{\leq{T_{h}}}},z^{n}\right)},\quad 1\leq{t}\leq{T_{p}},
            \label{eq:open-loop}
        \end{equation}
    \end{enumerate}
    \item The future trajectories are refined in a closed-loop fashion by resampling from a conditional distribution on surrounding map features and open-loop trajectories of other agents.
    \begin{equation}
        o_{T_{h}+t}^{n}\sim{
        p(o^{n}_{T_{h}+t}\mid\mathbf{S}_{T_{h}+t},
        \mathbf{M},b^{n})},\quad 1\leq{t}\leq{T_{p}},\ 1\leq{n}\leq{N}.
        \label{eq:closed-loop}
    \end{equation}
\end{enumerate}
Herein, modeling the arrival rate of the traffic agents is not the primary focus of this paper. We set $N$ to be independent of all other parameters during generation. In practice, the Poisson assumption of the number of agents in a scenario can be replaced by any distribution that best fits the data.

In the following sections, we introduce the key components to parameterize the above probabilistic model, including input representation for the map and trajectory features (Section~\ref{subsec: encoder}), the core decoder-only model $f$ that realizes the conditional distributions with adaptive-norm Transformer (Section~\ref{subsec: decoder}), and the algorithms for training (Section~\ref{subsec: training}) and sampling  (Section~\ref{subsec: sampling}) with variational Bayes.

\subsection{Input Representations}
\label{subsec: encoder}

As illustrated in Figure~\ref{fig: architecture}, we use three separate networks to encode the features representing map geometry, traffic signals, and trajectories. These networks project the concatenated low-dimensional raw information to a high-dimensional feature vector. Generally, they use embedding layers to encode discrete features, then concatenate with the rest of the continuous features, and finally project them to a unified high-dimensional space using a multi-layer perceptron (MLP).

\subsection{Decoder-only Transformer}
\label{subsec: decoder}

The core of NIVA is a decoder-only transformer that predicts the parameters in the conditional distributions. Specifically, it comprises a cascade of temporal self-attention blocks that parameterize the open-loop conditional prior and a cascade of adaptive-norm cross-attention blocks that parameterize the closed-loop distributions for resampling.

\paragraph{Relative Space-time Representation} For multi-agent simulation, one major challenge for applying attention is to find the appropriate reference frame for positional encoding. Previous works commonly adopt the \emph{focal-agent-centric} strategy, which normalizes all inputs to the coordinate system centered and aligned to the position and heading of an individual focal agent~\cite{Gao_2020_CVPR, NEURIPS2022_2ab47c96, Lu_2024_CVPR,pmlr-v238-lu24a}. Nevertheless, such projection is conducted on pairs of agents, which has a quadratic complexity concerning the number of agents, making it inefficient for multi-agent simulation. Therefore, we adopt the relative spacetime representation from recent works~\cite{Zhou_2023_CVPR,10398503,zhou2024behaviorgptsmartagentsimulation}. For query-key pair $(i,j)$, the positional encoding added to the key and value is given by
\begin{equation}
    R_{i,j}:=\text{MLP}\left(\gamma\left(\left\|d_{i,j}\right\|_{2},\angle\left(v_{i},d_{i,j}\right),\Delta\rho_{i,j},\Delta{t}_{i,j}\right) \right),
\end{equation}
where $d_{i,j}$ is the displacement vector pointing from object $i$ to $j$, $v_{i}$ is the orientation (\emph{i.e., angle of the displacement vector between consecutive steps}), $\angle\left(\cdot,\cdot\right)$ represent the difference in angle, $\Delta\rho_{i,j}$ is the difference in heading angle between $i$ and $j$, and $\Delta{t}_{i,j}$ is the time difference. We apply the Fourier feature mapping~\cite{NEURIPS2020_55053683} on the raw relative spacetime feature before passing it through an MLP. The Fourier feature mapping is given by
\begin{equation}
    \gamma(\mathbf{x}) := \text{Concat}[\cos(2\pi\mathbf{F}\mathbf{x});\sin(2\pi\mathbf{F}\mathbf{x})],
\end{equation}
where $\mathbf{F}\sim\mathcal{N}(0, \sigma^{2}\mathbf{I})$ is a transformation matrix initialized with elements as i.i.d samples from an isotropic Gaussian.

\paragraph{Temporal Self-Attention Blocks} The parameters in the conditional distribution of Eq.~\ref{eq:open-loop} are predicted by a cascade of temporal self-attention layers:
\begin{equation}
    \hat{\mu}_{s_{t}^{n}}, \hat{\sigma}^{2}_{s_{t}^{n}}\gets
    \text{MHSA}\Bigl(Q=z^{n}, K =o^{n}_{j} +R_{t,j},V =o^{n}_{j}+R_{t,j}\Bigr),
    \label{eq: temporal}
\end{equation}
where $j=1,\ldots,t-1$ and $\text{MHSA}$ is the multi-head self-attention layers, each consisting of layer normalization, multi-head attention, and residual connection, and feed-forward network~\cite{vaswani2023attentionneed}.

\paragraph{Adaptive-Norm Transformer Blocks} Using the mean from open-loop predictions from Eq~\ref{eq: temporal}, a cascade of multi-head cross-attention layers predicts in parameters of the conditional distribution in Eq.~\ref{eq:closed-loop}. Specifically, these layers update the prior means of each simulated agent by cross-attention to introduce interactive constraints from surrounding map elements and other agents.

To incorporate the behavior style $b^n$, we propose the adaptive-norm Transformer block, which is inspired by the DiT block in the Diffusion Transformer architecture~\cite{Peebles_2023_ICCV}. The block comprises a map-to-agent cross-attention layer, an agent-to-agent self-attention layer, and adaptive-norm operators to condition dynamics on styles and intentions. Similar to the temporal self-attention operation above, the map-to-agent and agent-to-agent cross-attention layer updates the current timestep feature by:
\begin{equation}
\begin{aligned}
    Q_{t,1}^{n} &\gets \alpha_{1}(b^{n})\odot
    \text{LayerNorm}(\hat{\mu}_{s_{t}^{n}})+\beta_{1}(b^{n}), \\
    Q_{t,2}^{n}&\gets\text{MHCA}\bigl(
    Q=Q_{t,1}^{n}, K=m_{j}+R_{n,j},V=m_{j} + R_{nj}\bigr), \\
    Q_{t,3}^{n}&\gets \delta_{1}(b^{n})\odot \\
    Q_{t,4}^{n}&\gets\alpha_{2}(b^{n})\odot
    \text{LayerNorm}(Q_{t,3}^{n})+\beta_{2}(b^{n}), \\
    Q_{t,5}^{n}&\gets\text{MHCA}\bigl(
    Q=Q_{t,4}^{n},K=\hat{\mu}_{s_{t}^{i}}+R_{ni},
    V=\hat{\mu}_{s_{t}^{i}}+R_{ni}), \\
    \hat{\mu}_{s_{t}^{n}}&\gets\delta_{2}(b^{n})\odot Q_{t,5}^{n}
\end{aligned}
\end{equation}
where $\odot$ is the element-wise multiplication, $m_{j}\in\mathbf{M}$ is the map feature to attend to, $i\neq{n}$ indicate surrounding agents, and $\text{MHCA}$ denotes multi-head cross attention. Herein, scale and shift factors $\alpha_{\cdot}$ and $\beta_{\cdot}$ are outputs from MLPs. In this way, we condition the attention on different behavior styles. The outputs from the decoder-only Transformer are the updated means and variances parameterizing the conditional distribution $p(s_{t}^{n}\mid{\mathbf{S}_{T_{h}+t},\mathbf{M},b^{n}})\sim\mathcal{N}(\mu_{s_{t}^{n}},\sigma^2_{s_{t}^{n}}\mathbf{I})$.

\subsection{Emission Model}
\label{subsec: emission}

To enable marginalization of high-dimensional latent state $s_{t}^{n}$, we leverage the property of conditional and marginal distributions of Gaussian distributions to build a linear emission model $p(o_{t}^{n}\mid{s_{t}^{n}})$, which yields closed-form marginal distribution for sampling and have a closed-form posterior distribution of it one would like to condition on observed $\mathbf{o}_{t}^{n}$. Specifically, we use a linear layer with weight $\mathbf{A}$. The marginal and conditional distributions are
\begin{equation}
    p(o_{t}^{n})\sim \mathcal{N}\left(\mathbf{A}\mu_{s_{t}^{n}}, \mathbf{A}(\sigma^{2}_{s_{t}^{n}}\mathbf{I})\mathbf{A}^{\intercal} + \varepsilon^{2}\mathbf{I}\right),
    \label{eq: predictive}
\end{equation}
\begin{equation}
    p(s_{t}^{n}|o_{t}^{n}) = \mathcal{N}\left(\mathbf{P}^{-1}\left[\mathbf{A}^{\intercal}(\varepsilon^{-2}\mathbf{I})o_{t}^{n}+(\sigma^{-2}_{s_{t}^{n}}\mathbf{I})\mu_{s_{t}^{n}}\right],\mathbf{P}\right),
\end{equation}
where $\mathbf{P}=\left(\sigma^{-2}_{s_{t}^{n}} + \mathbf{A}^{\intercal}\left(\varepsilon\mathbf{I}\right)\mathbf{A}\right)^{-1}$ and $\varepsilon$ is a small constant.

The closed-form marginal and conditional distributions enable us to sample current motion states $o_{t}^{n}$ directly from a marginal Gaussian distribution, and, conversely, update the mean of latent states $s_{t}^{n}$ given updated observation of the current state (see Section~\ref{subsec: sampling}).

\subsection{Training}
\label{subsec: training}

Despite having a closed-form posterior for $s_{t}^{n}$, the actual posterior for the style feature $b^{n}$ and the intention $z^{n}$ are intractable to compute in the hierarchical model. Therefore, we adopt the variational inference and approximate them by mean-field distributions:
\begin{equation}
    q(\mathbf{b},\mathbf{z};\phi) := \prod\limits_{n=1}^{N}q(b^{n};\phi)\cdot{q(z^{n};\phi)},
    \label{eq: recognition-model}
\end{equation}
where $q(b^{n};\phi)$ is a another Gaussian distribution. The mean and variance of $q(b^{n})$ are the outputs from a temporal-attention layer, aggregating all history and future motion states on the trajectory to the current timestep $t=T_{h}$.

In variational inference, we want to minimize the Kullback-Leibler (KL) divergence between the variational and true posterior. This is equivalent to maximizing the evidence lower bound of the log conditional likelihood:
\begin{equation}
    \begin{aligned}
        \mathcal{L}(\theta,\phi) := &
        -\sum\limits_{n=1}^{N}\sum\limits_{t=1}^{T}\mathbb{E}_{q(b^{n},z^{n})}\log{p}\left(o_{t}^{n}\right) \\
        &+ \sum\limits_{n=1}^{N}D_{\text{KL}}
        \left[q(z^{n};\phi)\left\|\right.p(z^{n})\right] \\
        &+ \sum\limits_{n=1}^{N}D_{\text{KL}}
        \left[q(b^{n};\phi)\left\|\right.p(b^{n})\right].
    \end{aligned}
    \label{eq: loss}
\end{equation}

\begin{table*}[!t]
    \caption{\textbf{Qualitative comparison to SOTA on Waymo Open Sim Agent Leaderboard.} Performance is evaluated on the testing dataset with $32$ rollouts. The best performance scores are highlighted in \textbf{bold}, while \underline{underline} indicates the second-best. $\downarrow$: Lower values are better. $\uparrow$: Higher values are better.}
    \centering
    \resizebox{\textwidth}{!}{%
    \begin{tabular}{@{}c|c|ccccc@{}}
    \toprule
    \textbf{Model Name} & \textbf{Realism}($\uparrow$) & \textbf{Kinematic}$(\uparrow$) & \textbf{Interactive}$(\uparrow)$ & \textbf{Map-based}($\uparrow$) & \textbf{minADE}($\downarrow$) & \textbf{\# Params} \\
    \midrule
    Linear Extrapolation~\cite{NEURIPS2023_b96ce67b} & 0.3985 & 0.2253 & 0.4327 & 0.4533 & 7.5148 & -     \\ 
    TrafficBotsV1.5~\cite{zhang2024trafficbots} & 0.6988 & 0.4304 & 0.7114 & 0.8360 & 1.8825 & 10M \\
    MVTE~\cite{wang2023multiversetransformer1stplace} & 0.7302 & 0.4503 & 0.7706 & 0.8381 & 1.6770 & - \\
    GUMP~\cite{hu2024solvingmotionplanningtasks}            & 0.7431 & 0.4780 & 0.7887 & 0.8359 & 1.6041 & -     \\
    BehaviorGPT~\cite{zhou2024behaviorgptsmartagentsimulation}     & 0.7473 & 0.4333 & 0.7997 & 0.8593 & \underline{1.4147} & 3M    \\
    KiGRAS~\cite{zhao2024kigraskinematicdrivengenerativemodel}          & \textbf{0.7597} & \underline{0.4691} & \textbf{0.8064} & \textbf{0.8658} & 1.4384 & 0.7M  \\
    \textbf{NIVA (Ours)}     & \underline{0.7543} & \textbf{0.4789} & \underline{0.8039} & \underline{0.8627} & \textbf{1.4112} & 1.0M  \\
    \bottomrule
    \end{tabular}}
    \label{tab: benchmark}
\end{table*}

\begin{algorithm}
    \caption{Training Algorithm for NIVA}
    \begin{algorithmic}
        \State $\theta, \phi \gets \text{Random Initialization}$
        \While{not converged}
            \State Sample a mini-batch $\left\{(\mathbf{O},\mathbf{M})\right\}_{1,\ldots,D}$
            \State $\mathcal{L}(\theta,\phi;D) \gets 0$
            \For{$d=1,\ldots,D$}
                \State Forward pass temporal self-attention layers
                \State Evaluate $q(z^{n}_{k})$ by Eq.~\ref{eq: log-qz}
                \State Select most consistent intention $k^{\ast}\gets\underset{k}{\text{argmax}}q(z_{k}^{n})$
                \State Calculate and sample $b^{n}\sim{q(b^{n};\phi)}$
                \State $\mathcal{L}(\theta,\phi;D) \gets \mathcal{L}(\theta,\phi;D)+\mathcal{L}(\theta,\phi;z_{k^{\ast}}^{n},b^{n})$
            \EndFor
            \State Calculate gradient $\mathbf{g}_{\phi} \gets \nabla_{\phi}L(\theta,\phi;D)$
            \State Calculate gradient $\mathbf{g}_{\theta} \gets \nabla_{\theta}L(\theta,\phi;D)$
            \State Update parameters $\theta,\phi\gets\text{optimizer}(\theta,\phi,\mathbf{g}_{\theta},\mathbf{g}_{\phi})$
        \EndWhile
    \end{algorithmic}
    \label{alg: training}
\end{algorithm}

The issue with this loss function is that the gradients for the variational parameters are functions of the ones of the generative distribution. To resolve the interleaved gradients, we apply the expectation-maximization (EM) algorithm to solve for the optimal parameter $\theta^{*}$. We first calculate the optimal variational distribution for intention assignment. The optimal variational distribution for $z^{n}$ is given by letting $\partial{L}/\partial{q(z^{n})}=0$, which yields a function about the open-loop conditional distribution
\begin{equation}
    q(z^{n}_{k})=\text{Softmax}\Bigl(
    \sum\limits_{t=1}^{T}
    \log{\int{p(o_{t}^{n}\mid{s_{t}^{n}})
    p(s_{t}^{n}\mid{\mathbf{O}_{\leq{T}_{h}},z^{n}_{k}}}})ds_{t}^{n}
    \Bigr).
    \label{eq: log-qz}
\end{equation}
The integration on the right-hand side yields a marginal linear Gaussian distribution. In the maximization step, we update the generative parameters $\theta$ by minimizing the loss function given in equation~(\ref{eq: loss}). The algorithm iterates between the two steps until convergence. The detailed algorithm is shown in Algorithm~\ref{alg: training}.

\subsection{Sampling}
\label{subsec: sampling}

To generate a traffic scenario from the NIVA model, we need to draw samples for all $n=1, \ldots, N$ agents. The benefit of the linear Gaussian model is that we can directly marginalize the latent variables $s_{t}^{n}$ to obtain the closed-form predictive distribution given by Equation (\ref{eq: predictive}). Suppose we aim to sample $L$ samples of a traffic scenario. The procedure is as shown in Algorithm~\ref{alg: sampling}.

\begin{algorithm}
    \caption{Scenario Sampling Algorithm for NIVA}
    \begin{algorithmic}
        \Require Prediction horizon $T_{p}$
        \Require Number of simulated agents $N$
        \Require Initial observations $\mathbf{O}_{1:T_{h}}$ and $M$
        \Require Prior parameter of intention distribution $\pi$
        \Require Trained NIVA model with parameters $(\theta,\phi)$
        \For{$n=1,\ldots,N$}
            \State Choose styles $b^{n}\sim\mathcal{N}(0,\mathbf{I})$
            \State Choose intention $z^{n}\sim\text{Cat}(\pi)$
            \State Sample variance vector $\nu_{n}\sim\mathcal{N}(0, \mathbf{I})$
            \For{$t=1,\ldots,T_{p}$}
                \State $\mu_{s_{t}^{n}},\sigma_{s_{t}^{n}}\gets\text{NIVA}(\mathbf{O}_{<T_{h}+t},\mathbf{M},z^{n},b^{n})$
                \State Compute cholesky decomposition $$R: \mathbf{A}(\sigma^{2}_{s_{t}^{n}}\mathbf{I})\mathbf{A}^{\intercal}+\varepsilon^{2}\mathbf{I}=RR^{\intercal}$$
                \State $o_{t}^{n} \gets A\mu_{s_{t}^{n}}+R^{\intercal}\nu_{n}$.
                \If{ground-truth observation $o_{t}^{n}$ is given}
                    \State Compute $P^{-1}=\Sigma_{s_{t}^{n}}^{-1} +A^{\intercal}A$
                    \State Update $\mu_{s_{t}^{n}}\gets \mathbf{P}^{-1}\left[\mathbf{A}^{\intercal}(\varepsilon^{-2}\mathbf{I})o_{t}^{n}+(\sigma^{-2}_{s_{t}^{n}}\mathbf{I})\mu_{s_{t}^{n}}\right]$
                \EndIf
            \EndFor
        \EndFor
    \end{algorithmic}
    \label{alg: sampling}
\end{algorithm}

%% file: sections/4_experiment.tex
\section{Experiments}
\label{sec: experiments}


\begin{figure*}
    \centering
    \includegraphics[width=\linewidth]{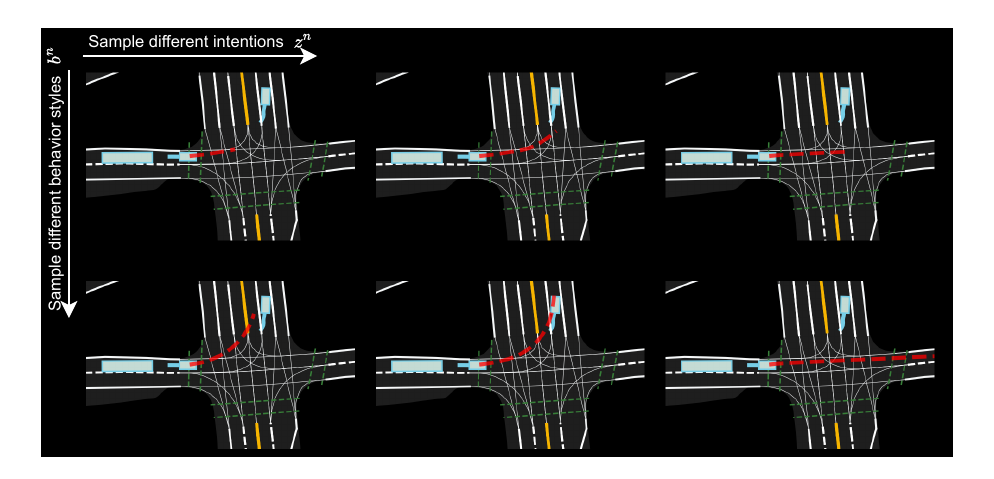}
    \caption{Sampling different behavior styles and intentions gives different future trajectories.}
    \label{fig: sample}
\end{figure*}

\subsection{Dataset}
\label{sec: dataset}
We train and evaluate NIVA using the Waymo Open Motion Dataset~\cite{Ettinger_2021_ICCV}. The dataset comprises $486,995$ training samples, $44,097$ validation samples, and $44,920$ testing samples, each with $T_{h}=1.1$ seconds history and $T_{p}=8$ seconds ground-truth trajectories collected at a frequency of $1$Hz. The dataset is the foundation for Waymo Sim Agents Challenge~\cite{NEURIPS2023_b96ce67b}, which is the first comprehensive benchmark with evaluation metrics on kinematic realism, interactive realism, and map compliance for autonomous driving simulation. The task requires the model to generate $32$ parallel simulations for each scenario. The prediction for each agent should contain the 3D coordinates of the centroid and heading angle of the bounding box at each future time step.

\subsection{Implementation Details}
\label{sec: implementation-details}
We encode agent and map features as $128$-dimensional vectors. For cross-attention, each agent only attends to $64$ nearest map points and five nearest agents. The intention codebook has a size of $K=6$. We train the model for $20$ epochs on four NVIDIA A100 GPUs with a batch size of $12$, using the AdamW optimizer~\cite{loshchilov2019decoupledweightdecayregularization} with weight decay $0.01$ and dropout probability $p=0.1$. We only apply weight decay regularization to the weights of the linear layers in the neural network. The learning rate increases from $0$ to $0.0002$ in the first $1000$ steps and decays to $0.0000003$ following a cosine annealing schedule~\cite{loshchilov2017sgdrstochasticgradientdescent}.

\begin{table}[!t]
\centering
\caption{\textbf{Effect of different nearby objects to attend in cross-attention.} Performance is evaluated on the validation dataset with $32$ rollouts.
}
\resizebox{\linewidth}{!}{%
\begin{tabular}{@{}ccccc@{}}
\toprule
$N_{\text{map}}$ &
$N_{\text{agent}}$ &
  \textbf{Realism} ($\uparrow)$ &
  \textbf{minADE} ($\downarrow$) &
  \textbf{Time (ms/scenario)}
\\ \midrule
64   &  2  & 0.7258 & 1.5128 &  28.2   \\
128  &  2  & 0.7371 & 1.4776 &  30.1   \\
256  &  2  & 0.7394 & 1.4728 &  37.4   \\
64   &  5  & 0.7472 & 1.4231 &  31.6   \\
128  &  5  & 0.7543 & 1.4112 &  35.7   \\
256  &  5  & 0.7543 & 1.4106 &  43.1   \\ \bottomrule
\end{tabular}}
\label{tab: neighborhood}
\end{table}

\begin{table}[!t]
\begin{minipage}{0.45\linewidth}
    \caption{\textbf{Performance with different ratios of training data.}}
        \centering
        \resizebox{\linewidth}{!}{%
        \begin{tabular}{@{}ccc@{}}
        \toprule
        \textbf{Ratio} & \textbf{Realism} & \textbf{minADE}\\ \midrule
        1\%            & 0.6285          & 1.8864          \\
        10\%           & 0.7028          & 1.6928          \\
        20\%           & 0.7412          & 1.6110          \\
        100\%          & 0.7594          & 1.4112          \\ \bottomrule
        \end{tabular}}
        \label{tab: scala-ratio}
\end{minipage}
\hfill
\begin{minipage}{0.49\linewidth}
    \caption{\textbf{Performance varying number of Transformer Blocks.}} 
        \centering
        \begin{tabular}{@{}cccc@{}}
        \toprule
        \textbf{$N_{\text{blocks}}$} &\textbf{Realism} & \textbf{minADE}\\ \midrule
        1    & 0.7543      & 1.4112          \\
        2    & 0.7574      & 1.4106          \\
        3    & --          & --              \\
        \bottomrule
        \end{tabular}
        \label{tab: scala-blocks}
\end{minipage}
\end{table}

\subsection{Realistic Traffic Generation}
\label{subsec: benchmark}
Table~\ref{tab: benchmark} shows evaluation metrics on the testing dataset. Notably, our proposed NIVA achieves the lowest minimum average displacement errors (minADE) and the best kinematic realism, highlighting the ability of our model to predict realistic dynamics of real-world agents. Meanwhile, it also achieves competitive performance compared to its counterpart models like KiGRAS~\cite{zhao2025kigras} and BehaviorGPT~\cite{zhou2024behaviorgptsmartagentsimulation}, revealing its capability to learn map constraints and interactions among nearby agents.

\subsection{Qualitative Analysis}
\label{subsec: qualitative}

To better showcase the generated scenario, Fig.~\ref{fig: sample} visualizes one of the simulated agents generated conditioned on the same history but with different style features and intention tokens. On the same row from left to right, the three intention tokens $z^{n}$ lead to three different destinations, where two are left-turning but into different exit lanes, while the other one is going straight. In the same column, the trajectories in the second row are longer and more aggressive than those in the first, reflecting a more aggressive behavior of the visualized agent. The results show that our proposed model can learn disentangled latent distributions that encode different behavior styles and intentions to sample from efficiently.

\subsection{Ablation Study}
\label{subsec: ablation}

In this section, we conduct ablation studies on several hyperparameters to better understand the proposed method.

\paragraph{Number of Attended Nearby Objects} Using the same trained model, we traverse different combinations of number of map objects $N_{\text{map}}$ and number of agents $N_{\text{agents}}$ in the neighborhood for cross-attention layers to investigate how it affect the final performance. As shown in Table~\ref{tab: neighborhood}, increasing the number of attended neighbor agents leads to a more significant performance improvement than increasing the number of attended map objects, highlighting the importance of constraints among interactive agents. Meanwhile, further increasing the number of map features or agents can increase inference time for generating a single scenario. It impacts the balance between performance and efficiency.

\paragraph{Scalability} Table~\ref{tab: scala-ratio} presents the performance of the proposed model trained on different proportions of the training dataset. According to the metrics, increasing the training data can significantly improve the performance, reflecting that our proposed model is scalable with data. In~\ref{tab: scala-blocks}, we compare performance with varying numbers of adaptive-norm Transformer blocks in the transition model. Despite improving the performance, increasing the number of blocks leads to a drastic increase in GPU memory, causing training failure when $N_{\text{blocks}}\geq 3$.

%% file: sections/5_conclusion.tex
\section{Conclusion}
\label{sec: conclusion}

This paper introduced Neural Interactive Agents (NIVA), a novel generative model for multi-agent traffic simulation. NIVA learns disentangled latent representation of continuous driving styles and discrete intention that are easy to sample. Experiments on the Waymo Open Dataset showed that NIVA achieves performance comparable to the state-of-the-art and can encode driving styles effectively. Future work will focus on improving interactive and map-based realism and extending NIVA to work with raw image or LIDAR point cloud inputs, enabling deployment in a real-world autonomous vehicle (AV) to further investigate the potential applications for the proposed framework. 